# Local Neighborhood Intensity Pattern – A new texture feature descriptor for image retrieval

[a]Prithaj Banerjee, [b]Ayan Kumar Bhunia, [c]Avirup Bhattacharyya, [d]Partha Pratim Roy*, [e]Subrahmanyam Murala

[a]Dept. of CSE, Institute of Engineering & Management, Kolkata, India. Email- [a]prithajtutanbanerjee@gmail.com
[b]Dept. of ECE, Institute of Engineering & Management, Kolkata, India. Email- [b]ayanbhunia007@gmail.com
[c]Dept. of ECE, Institute of Engineering & Management, Kolkata, India. Email- [c]avirupiem@gmail.com
[d]Dept. of CSE, Indian Institute of Technology Roorkee, India. Email- [d]proy.fcs@iitr.ac.in
[e]Dept. of EE, Indian Institute of Technology Ropar, India. Email- [e]subbumurala@iitrpr.ac.in
[d*]email: proy.fcs@iitr.ac.in, TEL: +91-1332-284816*

**Abstract**

In this paper, a new texture descriptor based on the local neighborhood intensity difference is proposed for content based image retrieval (CBIR). For computation of texture features like Local Binary Pattern (LBP), the center pixel in a 3×3 window of an image is compared with all the remaining neighbors, one pixel at a time to generate a binary bit pattern. It ignores the effect of the adjacent neighbors of a particular pixel for its binary encoding and also for texture description. The proposed method is based on the concept that neighbors of a particular pixel hold significant amount of texture information that can be considered for efficient texture representation for CBIR. The main impact of utilizing the mutual relationship among adjacent neighbors is that we do not rely on the sign of the intensity difference between central pixel and one of its neighbors ($I_i$) only, rather we take into account the sign of difference values between $I_i$ and its adjacent neighbors along with the central pixels and same set of neighbors of $I_i$. This makes our pattern more resistant to illumination changes. Moreover, most of the local patterns including LBP concentrates mainly on the sign information and thus ignores the magnitude. The magnitude information which plays an auxiliary role to supply complementary information of texture descriptor, is integrated in our approach by considering the mean of absolute deviation about each pixel $I_i$ from its adjacent neighbors. Taking this into account, we develop a new texture descriptor, named as Local Neighborhood Intensity Pattern (LNIP) which considers the relative intensity difference between a particular pixel and the center pixel by considering its adjacent neighbors and generate a sign and a magnitude pattern. Finally, the sign pattern ($LNIP_S$) and the magnitude pattern ($LNIP_M$) are concatenated into a single feature descriptor to generate a more effective feature descriptor. The proposed descriptor has been tested for image retrieval on four databases, including three texture image databases - Brodatz texture image database, MIT VisTex database and Salzburg texture database and one face database - AT&T face database. The precision and recall values observed on these databases are compared with some state-of-art local patterns. The proposed method showed a significant improvement over many other existing methods.

*Keywords*- Local Neighborhood Intensity Pattern, Local Binary Pattern, Feature Extraction, Texture Feature.

## 1. Introduction

An image is considered as an important source of information in many fields of research such as Computer Vision, Digital Image Processing, Pattern Recognition, etc. An image makes communication simpler as our brain can interpret them much quicker than text. The advancement of technology is overwhelming. Smart phones have almost become an integral part of our life. The increase in the use of such electronic gadgets has led to the evolution of digital images. Thus with the huge availability of digital images, gathering information from them also becomes essential and various problems like matching similar images or retrieving a blurred image has also become popular. Hence, image retrieval i.e. searching, browsing and retrieving similar images from a large collection of images has become a real challenge. In this context, Visual search engine and Reverse image search play pivotal roles. Our work is concerned with performing content based image retrieval by using the local structural and intensity information.

Manual annotation of images is both computationally and economically expensive. Hence, there is a dire need of expert automatic annotation and retrieval system which can work based on the content present in the image. In literature, researchers have focused on automatic image retrieval system (Jeena Jacob, Srinivasagan, & Jayapriya, 2014; Müller et al., 2004; Murala & Jonathan Wu, 2014; Murala, Maheshwari, & Balasubramanian, 2012b; MuralA & Wu, 2013; Subrahmanyam, Maheshwari, & Balasubramanian, 2012; Verma & Raman, 2015, 2017). Image searching and browsing based on metadata such as captions, keywords or any other text information depend on the availability of the text information. Content-based image retrieval (CBIR) is an important development in the field of computer vision in which an image is retrieved by avoiding the use of textual descriptions. Rather, it lays emphasis on similarities in their contents (textures, colors, shapes, etc.) with respect to the query image and lists down images from large databases in descending order of similarity. In the past few years, numerous researches have been done on content based image retrieval (a.W.M. Smeulders, Worring, Santini, Gupta, & Jain, 2000; Ahmadian, Mostafa, Abolhassani, & Salimpour, 2005; Celik & Tjahjadi, 2009; Dubey, Singh, & Singh, 2016; Felipe, Traina, & Traina, 2003; Guo, Zhang, & Zhang, 2010a; He, Sang, & Gao, 2012; Jeena Jacob et al., 2014; Shengcai Liao, Zhu, Lei, Zhang, & Li, 2007; Shu Liao & Chung, 2007; Murala et al., 2012b). This involves the process of extracting some features from the user defined query image and matching those features with the large image database. Again, different images have different features. Among the existing content-based descriptors for extracting features, texture is considered the most important as it refers to surface characteristics and appearance of an object given by the size, shape, density, arrangement, and proportion of its elementary parts.

The primary objective of developing a texture feature is to develop one that can work for a wide variety of images of different databases. Our work concentrates on this point and aims at giving equally effective results on different image datasets available online. The texture patterns proposed in our work compare the intensity of a particular pixel not only with the center pixel in a 3×3 window but also explore its relationship with other immediate adjacent neighboring pixels contained in the neighborhood for encoding it. Thus, our proposed method takes into account the effect of immediate adjacent neighboring pixels along with the center pixel in a 3×3 window for pattern calculation and proves to be more effective than the existing methods.

**1.1. Related work:**

Due to the fact that content based features such as texture, shape, and color are easy to extract, researchers have explored numerous methods for this purpose. Methods like i) Gabor filters (Ahmadian et al., 2005) ii) discrete wavelet transform (Loupias, Sebe, Bres, & Jolion, 2000) , etc. have been used for texture feature analysis. In Wavelet transform the signals are decomposed into a mutually orthogonal set of wavelets. Further improvements to this wavelet transform method was done in rotated wavelet filter(Kokare, Biswas, & Chatterji, 2007), rotated complex wavelet filter (Kokare, Biswas, & Chatterji, 2006) and dual tree complex wavelet transform (Celik & Tjahjadi, 2009) for texture features.

In the context of image retrieval, the most popular yet simple method is Local Binary Pattern (LBP) as proposed by Ojala et al.(T Ojala, Pietikainen, & Maenpaa, 2002). Besides its simplicity, the computational efficiency and robustness to illumination changes that it offers have greatly attracted the researchers. Application of LBP is manifold, be it in texture analyzing, medical image and motion analyzing, face detection or image searching and browsing. In LBP, a comparison is made between the intensity of the center pixel and its N neighboring pixels by considering them as lying on a circular neighborhood of radius R and the result is then encoded in binary form. This results in an N bit binary pattern. However, most of the works consider a radius of one and neighborhood of size 8. This results in an 8 bit binary pattern. In Fig. 1, illustration of the method is given to calculate the LBP value in a 3×3 neighborhood for R equals to 1 and N equals to 8. LBP has been chosen as a standard means to compare their results and further improvements were made on it in various methods like Completed LBP (Guo et al., 2010a), Dominant LBP (S Liao, Law, & Chung, 2009), Ellipse Topology (Shu Liao & Chung, 2007), Multi-scale Block Local Binary Pattern(MB-LBP) (Shengcai Liao et al., 2007), three and four batch LBP (Wolf, Hassner, & Taigman, 2008), Local Derivative Pattern(LDP) (B. Zhang, Gao, Zhao, & Liu, 2010). In fact, Gaussian as well as wavelet based low pass filters, were used in LBP called as Pyramid Local Binary Pattern(PLBP) (Qian, Hua, Chen, & Ke, 2011) as proposed by Qian et al. for texture classification. In this method, extraction of multi-resolution images from the original image is done by using a low pass filter, and the multi-resolution low pass images are used in LBP features collection. Again, a combined application of LBP along with the above mentioned Gabor filter gave better result (Tlig, Sayadi, & Fnaiech, 2012). Besides these, the concept of the moment was applied for feature extraction in (Papakostas, Koulouriotis, Karakasis, & Tourassis, 2013). Again, in Directional Local Extrema Pattern (Murala, Maheshwari, & Balasubramanian, 2012a), the feature was extracted by the edge information. Four directions in the image are utilised in this feature extraction by edge information method and further modification on it was done in (Subrahmanyam et al., 2012; Vijaya Bhaskar Reddy & Rama Mohan Reddy, 2014) which is then applied for image retrieval. Besides all these applications, there are other implementations which used the idea of LBP and made further modifications on it. Among these, Dominant Local Binary Pattern(DLBP) (S Liao et al., 2009), Local Bit-plane Decoded Pattern(LBDP) (Dubey et al., 2016), Local Edge Pattern for Segmentation and Image Retrieval (LEPSEG and LEPINV) (Yao & Chen, 2002), Multi-structure Local Binary Pattern (Ms-LBP) (He et al., 2012), Local Mesh Pattern (LMP) (Murala

& Wu, 2014), Average Local Binary Pattern(ALBP) (Hamouchene & Aouat, 2014) etc. are worth mentioning. Noise is a drawback of LBP. To minimize this effect various algorithms have been formulated. Again in Local Ternary Pattern (LTP) (Tan & Triggs, 2010), three bits -1, 0 and 1 are assigned to the neighboring pixels based on a threshold value with respect to the center. In this pattern firstly a threshold value is considered (say a), if the neighboring pixel values ($I_i$) is in the range of center pixel ($I_c$) $\pm$ threshold i.e. ($I_c - a, I_c + a$) then 0 is assigned and if it is less than this range -1 is assigned otherwise +1 is assigned. Then this ternary pattern is converted into upper and lower binary bit patterns. Its improved versions known as Improved LTP (Wu, Sun, Fan, & Wang, 2015) gives better result. Noise-Resistant LBP (NR-LBP) (Ren, Jiang, & Yuan, 2013), Robust LBP(RLBP) (Zhao, Jia, Hu, & Min, 2013) are also used for noise reduction of a normal LBP feature. In Local Tetra Patterns (Murala et al., 2012b) the second order derivation in horizontal and vertical directions are considered and it gives better result than LBP which is then transformed to binary patterns for calculations. Local Oppugnant Pattern (Jeena Jacob et al., 2014) is the extended version of Local Tetra Patterns in RGB color space. In spherical symmetric 3D Local Ternary Patterns(Murala & Jonathan Wu, 2015) proposed by Murala and Wu, Gaussian filters and RGB color space were used which provided a 3D space and extracted LTP from different directions .

Medical field is also benefitted by Content Based Image Retrieval in dealing with the large database since this helps in identifying diseases by retrieving similar images corresponding to a particular image. GNU Image finding tool has been used with histogram and Gabor filters (Müller et al., 2004) for this purpose. Local Mesh Patterns (LMeP) and Peak Valley Edge Pattern is also used for biomedical image retrieval (Murala & Wu, 2014; MuralA & Wu, 2013). These two methods are combined into Local Mesh Peak Valley Edge Patterns for CT and MRI image indexing and retrieval (Murala & Jonathan Wu, 2014).

Haralick used the concept of Gray Level Co-occurrence matrix (GLCM) for classification of the image by texture feature extraction. Here, the intensity of the pixels of images is considered based on some statistical features and mainly the number of occurrences of a particular kind of pattern is considered. In many cases, GLCM is used to calculate statistical parameters based on some operators such as Prewitt operator, operated on the image rather than operating on the real image (J. Zhang, Li, & He, 2008). Again in Center Symmetric Local Binary Co-occurrence Pattern (Verma & Raman, 2015), the feature is extracted based on the diagonally symmetric elements about the center. Moreover, in this proposed method GLCM is implemented for extracting the feature vector. In Local Neighborhood Difference Pattern(LNDP) (Verma & Raman, 2017), the feature is extracted by considering only the nearest element to a particular pixel and the final feature extraction for image retrieval is done by concatenating this LNDP feature with LBP feature. Different types of Local Descriptors in the literature along with their area of application are shown in Table 1.

To our knowledge, none of the previous methods considered the effect of the immediate adjacent neighbors of a particular pixel in a 3×3 window of an image for encoding it. However, the work in (Verma & Raman, 2016) considered only two neighbors of a particular pixel for its binary representation. In other words, the effect of neighboring pixels to calculate a binary pattern has not yet been fully explored. In this work, we explore the information contained in the adjacent neighbors of a particular pixel for encoding it rather than considering just two of its neighbors. Following this, we developed a novel method to calculate the binary pattern which takes into account both the sign of the intensity difference and magnitude of intensity difference with respect to adjacent neighbors. This is motivated by the fact that neighbors of a particular pixel also contain a significant amount of texture information.

The rest of the paper is organized as follows. In section 1.2, we have discussed the main contributions of our work and its advantages with respect to existing techniques. In section 2, the traditional LBP feature and the proposed pattern have been discussed. Section 3 describes the proposed system framework. Section 4 describes the datasets used in our work and the results obtained on those. The concluding part of the paper has been described in the last section.

Table 1: Different types of Local Descriptors in the literature

| Name of the Pattern | Area of Application | Reference |
|---|---|---|
| Local Binary Pattern(LBP) | Texture classification, face recognition, facial expression recognition, object tracking etc. | (Ahonen, Hadid, & Pietik$\\$$\"ainen, 2004; Ning, Zhang, Zhang, & Wu, 2009; Timo Ojala, Pietikäinen, & Harwood, 1996; Shan, Gong, & McOwan, 2009) |
| Local derivative pattern (LDP) | Facial Image Recognition | (B. Zhang et al., 2010) |
| LBP variance (LBPV) | Texture Classification | (Guo, Zhang, & Zhang, 2010b) |
| Pyramid based algorithm proposed using LBP and LBPV with Gaussian low pass filter | Smoke detection in Videos | (Yuan, 2011) |
| Local Maximum Edge Binary Pattern | Object Tracking | (Subrahmanyam et al., 2012) |
| Local Tri-directional Patterns | Texture descriptor for image retrieval | (Verma & Raman, 2016) |

## 1.2 Main Contribution:

A number of local patterns have been developed for image retrieval till now. However, most of the existing local patterns (Shengcai Liao et al., 2007; Moore & Bowden, 2011; Murala et al., 2012b; Nanni, Lumini, & Brahnam, 2010; B. Zhang et al., 2010) compare the intensity of a center pixel in a 3×3 window (see Fig. 1) with one of its 8 neighboring pixels at a time to encode it in binary form. Say for example, the popular LBP feature compares the intensity value between $I_c$ and $I_i$ at a time for encoding the binary pattern, where $I_c$ is the center pixel in a 3×3 window and $I_i$ is one of the 8 neighbors of $I_c$ (see Fig. 1). In other words, while comparing $I_c$ and $I_1$, the information in terms of the relative intensity difference between $I_1$ and its adjacent neighbors ($I_2, I_3, I_7, I_8$) in that 3×3 neighborhood are not considered. The idea of adjacent neighbors with respect to a particular pixel ($I_i$) is explained in Fig. 2. A similar case happens while comparing $I_c$ and $I_i$, for i = {2, 3, 4 ... 8}. In Center Symmetric LBP (CSLBP), the difference of the center symmetric pixels is used for calculating the local pattern of the input image without considering the remaining neighbors. In Local Ternary Pattern(Tan & Triggs, 2010), three bits -1,0 and 1 are assigned based on the comparisons made between the center and one of its neighboring pixels at a time and then this ternary value is converted to two binary patterns. Thus, in all these above methods we observe the information is obtained based on the encoding done by a direct comparison of the center pixel and one of its neighbors at a time. This encoding technique does not take in account the local information contained in the neighborhood as it concentrates only on one neighbor at a time and ignores the effect of others. In contrast, our



method is novel in the sense that it explores contribution of adjacent neighboring pixels for calculation of the binary pattern. This has motivated us to develop two binary patterns proposed in this paper which center around the idea of capturing the information contained in the adjacent neighborhood of a pixel for pattern calculation.

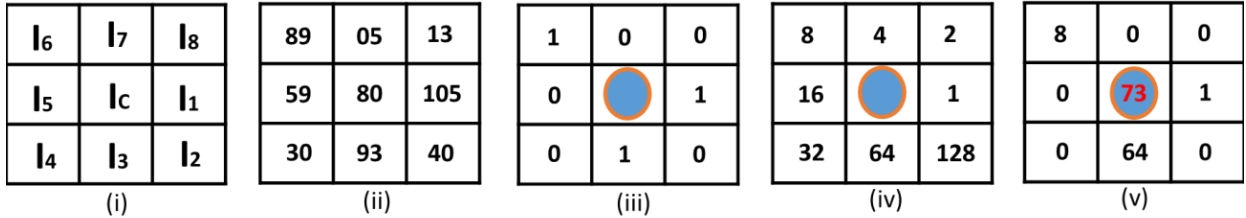

**Fig.1. Example showing the calculation of Local Binary Pattern (LBP) (i) a 3×3 window with general notations of the center and its neighboring elements (ii) an example of a 3×3 window (iii) threshold neighboring pixels with center and binary values are assigned accordingly (iv) specific weights (v) multiplying with weights and adding up to replace the result in the center.**

The first pattern compares a particular neighbor ($I_i$) with its adjacent neighbors(See Fig. 2) in a 3×3 window to generate a M-bit binary pattern which is compared to another M-bit pattern obtained by comparing center pixel with the same set of neighboring pixels in that 3×3 neighborhood. These two M-bit binary patterns are compared to efficiently encode their structural difference obtained by calculating the number of positions in which their respective bits differ to generate a single bit for $I_i$. Here, M = 4, if i ∈ odd number; M = 2, if i ∈ even number. We refer to section 2.2 for more details about the calculation. Similarly, we generate an 8-bit pattern by performing the same method for each of the 8 neighbors ($I_i \forall i = 1,2,....8.$) of the center pixel which is replaced at the center ($I_c$) by the decimal equivalent of the obtained 8-bit pattern for that 3×3 window. Similarly, our second pattern computes the relative intensity value for a particular neighbor ($I_i$) based on the mean of absolute deviation about a particular pixel ($I_i$) from its adjacent neighbors in that 3×3 window. A threshold value ($T_c$) is calculated by taking the mean deviation of the neighbors ($I_i$) about the center pixel ($I_c$) and then the comparison is made with this threshold value $T_c$ to evaluate the bit pattern. We concatenate the histogram of the two proposed patterns to optimize the image retrieval performance, which are collectively named as Local Neighborhood Intensity Pattern (LNIP) since both of these consider the contribution of the Local Neighborhood. However, the first pattern considers the sign (S) of the intensity difference between a particular pixel ($I_i$) and its adjacent neighbors in a 3×3 window and thus is named $LNIP_S$. The second pattern is named $LNIP_M$ as it considers the



absolute difference (M) of pixel intensity values. The most important contribution of our paper is that we exploit the relation that a particular pixel holds with its adjacent neighboring pixels in a 3×3 neighborhood.

Our proposed pattern compares influence of adjacent neighboring pixels for calculating the bit pattern. Thus, it extracts more discriminative information necessary for retrieval purposes. As we are encoding the relative intensity difference or relative sign changes with respect to adjacent neighborhood, our feature descriptor is less sensitive to gray level changes and provides robustness against illumination.

In brief, the salient contributions of our work may be stated as follows: (a) we develop two new texture patterns for expert image retrieval system named $LNIP_S$ and $LNIP_M$. $LNIP_S$ focuses on the sign of the relative intensity difference between the pixels. (b) The $LNIP_M$ captures the information in terms of mean deviation about a particular pixel and its adjacent neighbors with respect to the threshold value $T_c$ to calculate the binary pattern. (c) The resultant feature descriptor has been used for image retrieval on different databases (Brodatz texture image database, MIT VisTex database, Salzburg texture database and AT&T face database) and it has yielded promising results by performing better than the popular LBP feature and showing a competitive performance with other recent texture descriptors.

## 2. Local Pattern

### 2.1 Local Binary Patterns

As proposed by Ojala et al., Local Binary Pattern(LBP) was mainly invented for texture classification(T Ojala et al., 2002). Due to its computational ease, it was further used in object tracking(Ning et al., 2009), facial expression recognition (Moore & Bowden, 2011), medical imaging (Nanni et al., 2010) and image classification (T Ojala et al., 2002). In this method, a small window of an image is considered and the intensity difference between the center pixel and its N neighbors lying on the circumference of a circle of radius R is encoded in a binary format. Based on the intensity difference, a binary value (0 or 1) is assigned to each of the surrounding pixels (as given in eqn.1) which again multiplied by some specific weights sum up to give the result. This value is replaced with the value of the center pixel which is the binary pattern value



for that center pixel. This has been explained diagrammatically in Fig. 1. Thus, a local binary map of the image is generated in its gray level by replacing each center pixel with its binary pattern value in a similar manner. The feature vector is calculated by creating a histogram of this local binary map. The formula for LBP and the histogram is defined in eqn. (1)-(4).

$$LBP(N, R) = \sum_{i=1}^{N} 2^{i-1} \times D_1(I_i, I_c) \quad \ldots \ldots \ldots (1)$$

$$D_1(I_i, I_c) = \begin{cases} 1, & I_i \geq I_c \\ 0, & \text{otherwise} \end{cases} \quad \ldots \ldots \ldots (2)$$

$$His(N)|_{LBP} = \sum_{i=1}^{X} \sum_{j=1}^{Y} D_2(LBP(i,j), L); \quad L \in [0, (2^N - 1)] \quad \ldots \ldots \ldots (3)$$

$$D_2(a_1, a_2) = \begin{cases} 1, & a_1 = a_2 \\ 0, & \text{otherwise} \end{cases} \quad \ldots \ldots \ldots (4)$$

Here, N and R represent the number of neighboring pixels and radius respectively. $I_i$ denotes the $i^{th}$ surrounding pixel and $I_c$ denotes center pixel. Eqn. (3) is used to compute the final histogram of the pattern map. An example of LBP calculation is shown in Fig. 1.

**2.2 Local Neighborhood Intensity Pattern**

Local Neighborhood Intensity Pattern (LNIP) is an extension of popular LBP. In addition, the other most similar work related to our method is (Verma & Raman, 2016) which also considers two neighboring pixels for binary pattern calculation. However, there are several remarkable differences between our method and (Verma & Raman, 2016) which will be described in this section. We consider a radius of unit distance since closest neighboring pixels holds more discriminating information for texture descriptors. Thus, we use a 3×3 window to calculate our proposed binary pattern. Our proposed pattern intends to explore the mutual information with respect to the adjacent neighbors. The definition of adjacent neighbors in a 3×3 window is provided using a diagram in Fig. 2. Consider a center pixel $I_c$ and its 8 neighborhood pixels $I_1, I_2, \ldots \ldots, I_8$. From the Fig. 2, it can be inferred that $I_i$ among the 8 neighborhood pixel in a 3×3 window has 4 adjacent neighboring pixels if i ∈ odd; and 2 adjacent neighboring pixels if



i ∈ even. We denote the set of adjacent neighboring pixels with respect to $I_i$ as $S_i$. So, the number of elements in $S_i$ is 4 and 2 for i ∈ odd and i ∈ even respectively. Mathematical definition of $S_i$ is given by eqns. (5)-(6).

$$S_i = \{I_{1+\mod(i+5,7)}, I_{1+\mod(i+6,9)}, I_{i+1}, I_{\mod(i+2,8)}\} \ \forall \ i = 1,3,5,7 \quad \ldots\ldots\ldots (5)$$

$$S_i = \{I_{i-1}, I_{\mod(i+1,8)}\} \ \forall \ i = 2,4,6,8 \quad \ldots\ldots\ldots (6)$$

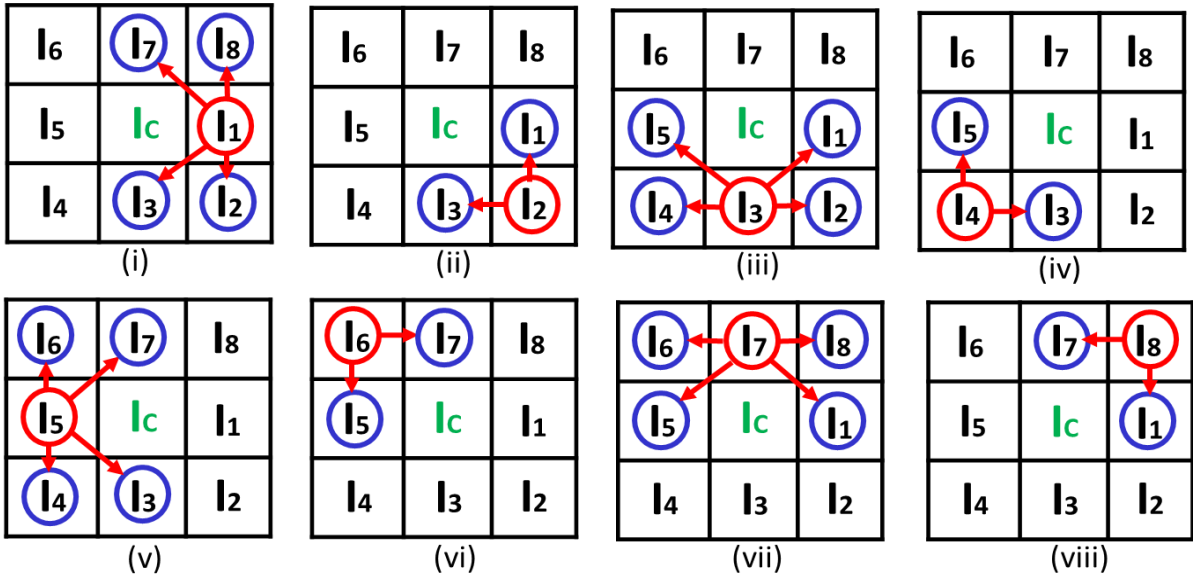

**Fig. 2** Diagram showing the adjacent neighbor relation for each of the 8 neighbors ($I_i \ \forall \ i = 1, 2, \ldots 8.$) of pixel $I_c$. $I_i$ has 4 adjacent neighbors if i = 1, 3, 5, 7; i.e. i ∈ odd. $I_i$ has 2 adjacent neighbors if i = 2, 4, 6, 8; i.e. i ∈ even. The set of adjacent neighbors of $I_i$ is denoted by $S_i$.

For our signed pattern ($LNIP_S$), we first calculate the sign of the relative difference between one of the 8 neighbors of center pixel ($I_i$) and its corresponding adjacent neighbors $S_i$. We get an M bit pattern with respect to $I_i$ where M is the number of elements in $S_i$. We get a binary pattern $B_{1,i}$ corresponding to $I_i$ using eqn. (7). Similarly, we calculate another binary pattern $B_{2,i}$ of same size corresponding to the center pixel considering the same neighboring pixels as in $S_i$ using eqn. (8).

$$B_{1,i}(k) = \text{Sign}(S_i(k), I_i) \text{ where, } k = 1 \text{ to } M. \ldots\ldots\ldots (7)$$

$$B_{2,i}(k) = \text{Sign}(S_i(k), I_c) \text{ where, } k = 1 \text{ to } M. \ldots\ldots\ldots (8)$$

$$\text{Sign}(a, b) = \begin{cases} 1, & a \geq b \\ 0, & \text{otherwise} \end{cases} \quad \ldots\ldots\ldots (9)$$



Here, a and b are two numbers. Following these, the single bit with respect to $I_i$ is evaluated simply by comparing the change in the structure of these two binary patterns $B_{1,i}$ and $B_{2,i}$. The structural change in the bit pattern is calculated by taking bitwise XOR operation between these two patterns. XOR between two same bits gives 0 as output and XOR between two dissimilar bits gives 1 as output. So XOR(1,0) and XOR(0,1) gives 1 as output whereas XOR(0,0) and XOR(1,1) gives 0 as output. On calculating bitwise XOR an M bit pattern $D_i$ is obtained as in eqn. (10). Since XOR of two dissimilar bits is assigned with 1 hence the total count of 1 in $D_i$ given by $\#(D_i = 1)$ gives the total structural change in $B_{2,i}$ compared to $B_{1,i}$ bit pattern. For two M bit patterns the total number of positions at which the respective bits may differ ranges from 0 to M. Here $\frac{1}{2}(M)$ is considered as threshold where, M = 4, if i ∈ odd number; M = 2, if i ∈ even number A binary bit for $I_i$ is defined based on a threshold as given by eqn. (11). Similarly, we calculate every pattern value corresponding to 8 neighboring pixels of $I_c$ in a 3×3 window using $I_i$ and its $S_i$, where i = 1, 2, 3….,8. The final value corresponding to central pixel ($I_c$) is calculated using eqn. (12) which is replaced in place of it at the final binary pattern map. After getting this pattern map, the histogram is calculated using eqn. (3). More explanation of our pattern value calculation for $LNIP_S$ using example is given in Fig. 4.

$$D_i = XOR(B_{1,i}, B_{2,i}) \quad \ldots \ldots \ldots (10)$$

$$B(I_i, I_c) = \begin{cases} 1, & \#(D_i = 1) \geq \frac{1}{2}(M) \\ 0, & \text{otherwise} \end{cases} \quad \ldots \ldots \ldots (11)$$

$$LNIP_S(I_c) = \sum_{i=1}^{8} 2^{i-1} \times B(I_i, I_c) \quad \ldots \ldots \ldots (12)$$

In various works (Murala et al., 2012b; Pan, Li, Fan, & Wu, 2017; Verma & Raman, 2016; Zhao, Huang, & Jia, 2012) it has been found that magnitude pattern is also helpful to create more informative feature descriptor. However, the work in (Zhao et al., 2012) clearly pointed out that sign vector of the difference vector preserves more discriminating information compared to its magnitude part. In local patterns like LBP only sign pattern is considered. So, while assigning



binary bits only the sign of the difference between each neighboring pixel and center pixel is considered. LBP does not consider the magnitude of these differences. As shown in the Fig.3, different local structural patterns can be encoded to the same LBP code which is not appropriate.

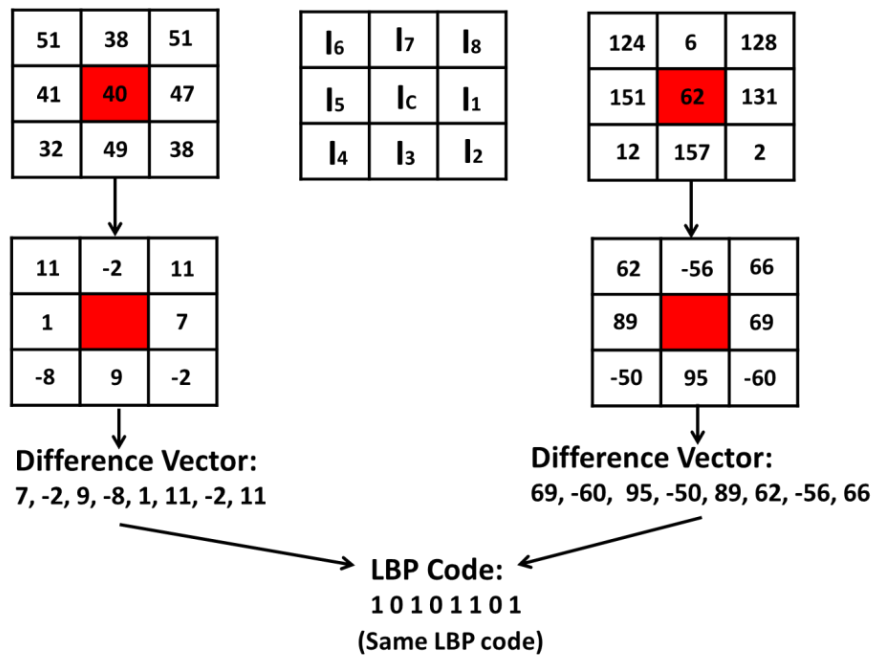

**Fig. 3. An example showing that two completely different local structural patterns can be coded inappropriately to the same LBP code.**

In our proposed pattern, we consider a magnitude pattern along with sign pattern. The concept of statistical dispersion is implemented to calculate our magnitude pattern for a 3×3 window. In order to calculate this dispersion, absolute mean deviation about a particular pixel is considered and this is compared with the mean deviation about the center pixel. Note that, normal average of all the differences will not work as the positive and negative terms will cancel each other to give a general average. Hence, we focused more to find out the total dispersion of the adjacent neighboring pixels from a particular pixel. For this purpose, the absolute mean deviation is considered as our statistical tool. We calculate the magnitude pattern which encodes the mean deviation of the adjacent neighbors ($S_i$) about a particular pixel ($I_i \forall\ i = 1, 2, ... 8$) with respect to a threshold value ($T_c$). The threshold value ($T_c$) is calculated by taking the mean deviation of the neighbors ($I_i$) about the center pixel ($I_c$). Thus, in contrast to (Verma & Raman, 2016) we have modified the formulation for magnitude pattern in order to take advantage of both the sign and



magnitude difference vector. We consider the mean of absolute deviation of adjacent neighbors to acquire the information of total deviation of these neighbors from a particular pixel ($I_i$) to calculate the magnitude pattern. $LNIP_M$ is calculated as follows.

$$M_i = \frac{1}{M} \sum_{k=1}^{M} | S_i(k) - I_i | \quad \ldots\ldots\ldots (13)$$

$$T_c = \frac{1}{8} \sum_{i=1}^{8} | I_i - I_c | \quad \ldots\ldots\ldots (14)$$

$M_i$ is the mean deviation about $i^{th}$ neighbor of $I_c$ from its corresponding $S_i$ (where i = 1, 2, 3….,8 following the eqn. (13) ) calculated for all the 8 neighboring pixels of $I_c$ in a 3×3 window. Here, $T_c$ is the threshold value which is calculated by taking the mean deviation of the neighbors ($I_i$) about the center pixel ($I_c$) by eqn. (14). Following this, the final bit for $i^{th}$ neighbor of $I_c$ is calculated by comparing $M_i$ with the threshold value $T_c$ using eqn. (15). The final magnitude pattern value corresponding to $I_c$ is evaluated using eqn. (16).

$$M(I_i, T_c) = \text{Sign}(M_i, T_c) \quad \ldots\ldots\ldots (15)$$

$$LNIP_M(I_c) = \sum_{i=1}^{8} 2^{i-1} \times M(I_i, T_c) \ldots\ldots\ldots (16)$$

Similarly, the histogram of our magnitude pattern is calculated using eqn. (3) and two final histograms of our Local neighborhood Intensity Pattern, i.e. $LNIP_S$ and $LNIP_M$, are concatenated to form the final feature descriptor as in eqn. (17).

$$\text{Hist} = [\text{Hist}^{LNIP_S}, \text{Hist}^{LNIP_M}] \quad \ldots\ldots\ldots (17)$$



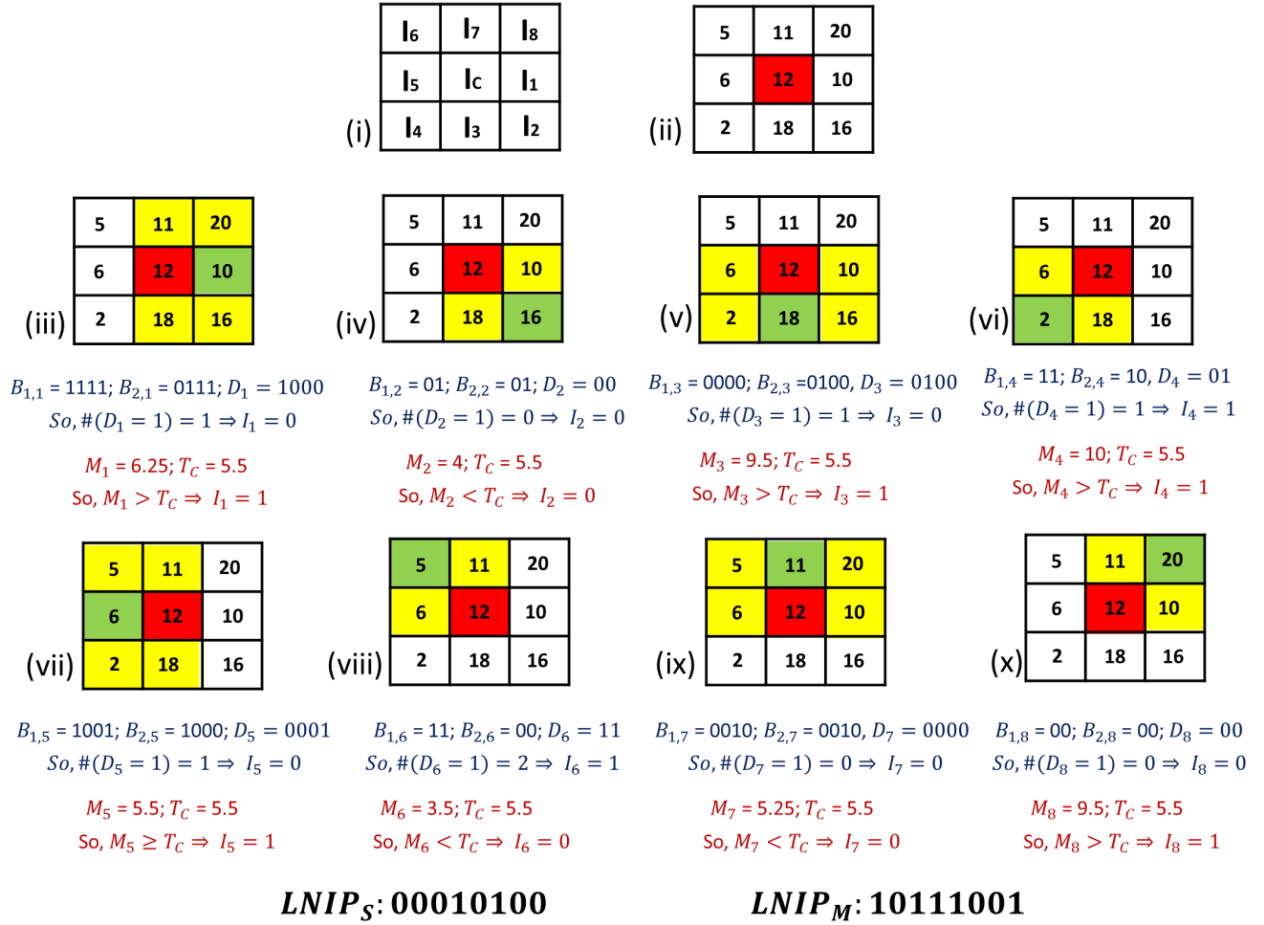

**Fig. 4** Example showing the calculation of General Local Neighborhood Intensity Pattern (i) a 3×3 window with general notations of the center and its neighboring elements (ii) an example of a 3×3 matrix window (iii)-(x) Local Neighborhood Intensity sign and magnitude pattern calculation.

Our proposed sign and magnitude pattern is calculated with an example as shown in Fig. 4 through (i)-(x) windows. Center pixel $I_c$ and neighborhood pixels $I_1, I_2, I_3 \ldots, I_8$ are shown in window (i) with center pixel marked with red color. In Fig. 4(ii) an example window is shown. In Fig. 4, windows ((iii)-(x)) calculation of our pattern is shown with the example in Fig. 4(ii). Here, center pixel is marked with red, pixel in concern is shown with green color and its adjacent neighbors are marked with yellow color. In Fig. 4(iii), first neighboring pixel $I_1$, is marked with green color, and its four adjacent pixels marked with yellow color. First, we compare green pixel with a set of yellow pixels and assign '0' or '1' value for all the four comparisons and similarly we compare red pixel with the same set of yellow pixels and assign '0' or '1'. As in Fig. 4(iii) $I_1$ and $I_c$ are compared with $I_2, I_3, I_7$ and $I_8$. Since $I_1$ is lesser than $I_2, I_3, I_7$ and $I_8$ we get a 4-bit pattern 1111 and since $I_c > I_7$ and $I_c$ is lesser than $I_2, I_3$ and $I_8$ so only $I_7$ bit is 0 and the other 3-



bits are 1 thus we obtain a 4-bit pattern as 0111. An exclusive-or (XOR) operation is then performed bitwise between the individual bits of the two binary strings 1111 and 0111. This results in a 4 bit string 1000. The number of positions in which the respective bits differ for the two strings ($\#(D_1 = 1)$) is 1 in this case. This is less than the threshold value of 2 (for i $\in$ odd). Hence, the bit designated for $I_1$ is 0. In a similar way, for next set of windows (iv)–(x), pattern values are obtained using corresponding neighboring pixels. For pixels having odd number of neighbors, for example $I_4$, the two bit pattern obtained is 11 while the two bit pattern obtained for the center pixel is 10. The XOR operation between the individual bits produces the binary string 01.The number of transitions in this case ($\#(D_4 = 1)$) is 1 which is equal to the threshold value 2(for i $\in$ even). Thus, the bit value obtained for $I_4$ is 1. The bits for the remaining pixels are calculated in a similar manner. Finally, our sign pattern for center pixel is calculated by combining all neighborhood pixel pattern values and representing them as a binary string.

Magnitude pattern is calculated for the same set of windows (iii)-(x). Mean deviation of the adjacent neighborhood pixels about a particular pixel ($I_i$) is obtained and these two values are compared with the threshold value $T_c$ as calculated in eqn. (14). In our example, 12 is center pixel and $I_1$ is 10. In window (iii), mean deviation of the adjacent neighborhood pixels about $I_1$ is calculated as shown in eqn. (13) comes out to be 6.5 and the threshold value $T_c$ is 5.5. Since, the calculated magnitude pattern value of $I_1$ is greater than the threshold value $T_c$, we assign 1 pattern value here. Similarly, magnitude pattern is calculated for next neighborhood pixels and shown in (iii)–(x) windows, and magnitude bits of all neighborhood pixels are represented in the form of a binary string and that is magnitude pattern for center pixel.

## 2.3 Advantages over other methods

The purpose of local patterns is to gather information from neighboring pixels in order to extracts features based on local intensity variation. In case of original LBP, comparisons are made between the center pixel and its one of the neighboring pixels at a time to encode a particular bit value. In contrast, the texture pattern proposed in our work utilizes the relation with adjacent neighboring pixels to calculate the binary pattern. In most of the existing works, the mutual relationship among the neighbors of center pixel is not considered. The main advantages of utilizing the mutual relationship among adjacent neighbors are as follows: Firstly, it makes the



pattern more resistant to illumination changes and the mutual relationship among the adjacent neighbors is well explored in this pattern. Secondly, we do not rely on the sign of the intensity difference between central pixel and one of its neighbors ($I_i$) only, rather we take into account the sign of difference values between $I_i$ and its adjacent neighbors along with the central pixels and same set of neighbors of $I_i$. Then the change in the structure is encoded in to binary bits. This makes our pattern more resistant to illumination changes and the information of the variation of neighbor pixel's structure from the center pixel is also gathered. This variation in structure is encoded into 0's and 1's which helps to define a more prominent relationship between the center pixel and its neighbors. Thirdly, most of the local patterns including LBP concentrates mainly on the sign information and thus ignores the magnitude. The sign information is more important for any binary pattern as mentioned in (Guo et al., 2010a), however, the magnitude information cannot not be eliminated since it plays an auxiliary role to supply complementary information texture descriptor. So in our magnitude pattern we considered the mean of absolute deviation about each pixel $I_i$ from its adjacent neighbors ($S_i$). Both our sign pattern ($LNIP_S$) and magnitude pattern ($LNIP_M$) poses complementary information and thus upon concatenation it gives superior performance for image retrieval task.

## 3. Proposed System Framework

A block diagram of the proposed method is presented in Fig. 5, and algorithm for the same is given below. The algorithm is given in two parts. Part 1 describes the system framework. Here, an image is taken as input and in the output, the feature vector is obtained after concatenating the histograms generated after applying $LNIP_S$ and $LNIP_M$. In part 2, retrieval of image is performed using $LNIP_S$ and $LNIP_M$. Here, query image is taken as input and in output the retrieved images are obtained based on similarity measure of the feature vectors as in part 1.



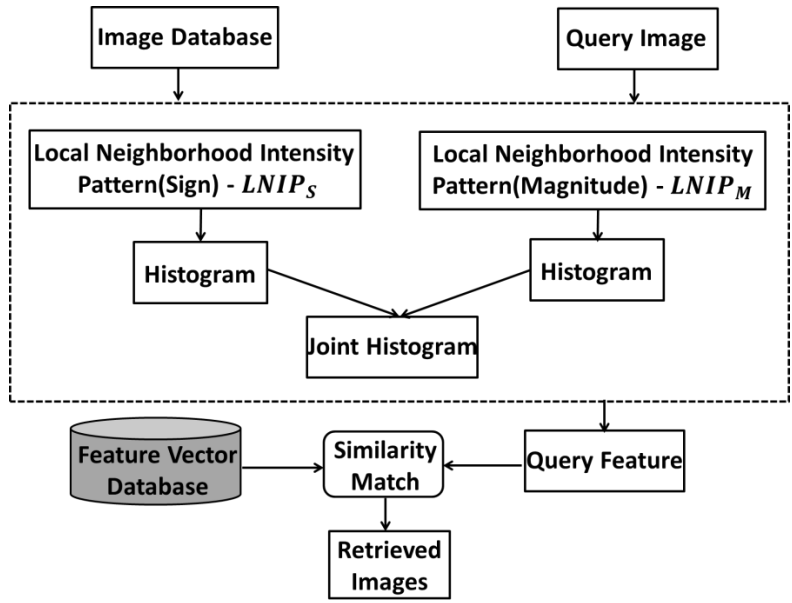

**Fig. 5. Proposed system block diagram.**

### 3.1.3 System Framework Algorithm:

**Part 1: Feature vector construction**

**Input: An Image from the database.**

**Output: Feature vector.**

1) Take an image from the image database and convert it into gray scale image if the image is colored.

2) Apply $LNIP_S$ and compute the histogram.

3) Apply $LNIP_M$ and compute the histogram.

4) Concatenate the two histograms obtained in step-2 and step-3 to obtain the feature vector.

**Part 2: Image retrieval using $LNIP_S$ and $LNIP_M$**

**Input: Query image from the database**

**Output: Retrieved images based on similarity measure**

5) Take the query image as input.

6) Perform step-2 to step-3 in part 1 to extract the feature vector of the query image.

7) Perform the similarity measure to compute the similarity index of the query image vector with every database images using different similarity measures.

8) Sort the similarity indices from highest to lowest to get the set of similar images.

9) Evaluate the performance using the metrics



## 3.2. Similarity Measure:

To retrieve the images in a content based image retrieval technique, alongside feature vector calculation, similarity measure also plays an important role. After calculating the feature vectors, this similarity measure gives the distance between the query image feature vector and feature of every image from the database i.e., the dissimilarity between the images is found out. Based on this measure indexing is done and indices with lower measures are sorted out as the set of retrieved images. Following five distance measures are used for calculation of similarity matching.

a. d1 distance:

$$DM_{D,q_i} = \sum_{j=1}^{n} \left| \frac{F_d^i(j) - F_{q_i}(j)}{1 + F_d^i(j) + F_{q_i}(j)} \right| \quad \ldots\ldots\ldots (18)$$

b. Euclidean Distance

$$DM_{D,q_i} = \left( \sum_{j=1}^{n} |(F_d^i(j) - F_{q_i}(j))^2| \right)^{1/2} \quad \ldots\ldots\ldots (19)$$

c. Manhattan Distance

$$DM_{D,q_i} = \sum_{j=1}^{n} |F_d^i(j) - F_{q_i}(j)| \quad \ldots\ldots\ldots (20)$$

d. Canberra Distance

$$DM_{D,q_i} = \sum_{j=1}^{n} \left| \frac{F_d^i(j) - F_{q_i}(j)}{F_d^i(j) + F_{q_i}(j)} \right| \quad \ldots\ldots\ldots (21)$$

e. Chi-square Distance



$$DM_{D,q_i} = \frac{1}{2}\sum_{j=1}^{n}\frac{(F_d^i(j) - F_{q_i}(j))^2}{F_d^i(j) + F_{q_i}(j)} \quad \ldots\ldots\ldots (22)$$

Here, $DM_{D, q_i}$ represents the distance function for database $D$ and query image $q_i$, n represents the length of the feature vector. $F_d^i(j)$ and $F_{q_i}(j)$ are feature vector of $i^{th}$ database image and query image respectively.

## 4. Experimental Results and Analysis
### 4.1 Evaluation Protocol

To validate the performance of the proposed method, it has been tested on three texture databases and one face image database for image retrieval. The performance of the proposed method for image retrieval is evaluated on the basis of precision and recall. The superiority of the proposed method has been verified by comparing it with some recent texture patterns for image retrieval in terms of the evaluation metrics mentioned earlier. For each database, some images have been retrieved based on a query image. Each and every image of the database has been treated as a query image once for each database. Before proceeding further, the concept of relevant and non-relevant images should be clearly understood. Relevant images are those that belong to the same category as that of a particular query image. The remaining images in the database may be treated as non-relevant images for that query image. The precision is a ratio of the total number of relevant images retrieved from the database to the total number of images retrieved from the database. In other words, precision is also known as positive predictive value and it is a function of the true positive and false positive images in the retrieval system. Recall is the ratio of the total number of relevant images in retrieved images to the total no of relevant images in the database. Recall is thus a function of the true positive images and false negative images in the retrieval system. If the total no of images retrieved is N, then the precision and recall may be computed as:

$$\text{Precision Rate}(P_r) = \frac{\text{Total no of relevant images retrieved from the database}}{\text{Total no of images in the databases}(N)}$$



$$\text{Recall Rate}(R_r) = \frac{\text{Total no of relevant images retrieved from the database}}{\text{Total no of relevant images present in the database}}$$

The average precision rate may be calculated as follows:

$$P_{avg}(C) = \frac{1}{l}\sum_{r=1}^{l} P_r \ldots\ldots\ldots (23)$$

In eqn. 23 $P_{avg}(C)$ represents the average precision rate for category (C), where l is the total no of images in that category. Similarly, the recall rate for each category may be expressed as given in eqn. 24

$$R_{avg}(C) = \frac{1}{l}\sum_{r=1}^{l} R_r \ldots\ldots\ldots (24)$$

On similar terms, we can compute the total precision and total recall for our experiment using eqn. 25 and 26.

$$P_{total}(C) = \frac{1}{K}\sum_{c=1}^{K} P_{avg}(C) \ldots\ldots\ldots (25)$$

$$R_{total}(C) = \frac{1}{K}\sum_{c=1}^{K} R_{avg}(C) \ldots\ldots\ldots (26)$$

Here, K is the total no of categories present in the database. Total Recall is also known as Average Recall Rate (ARR). The performance of the proposed method has been compared with a number of state of the art methods. The list of abbreviations for these methods has been given in Table 2.

Table 2: - Abbreviations for the list of methods used:

| Method | Name |
|---|---|
| CSLBP | Center Symmetric local binary pattern(Heikkilä, Pietikäinen, & Schmid, 2006) |
| LEPINV | Local Edge Pattern for Image Retrieval(Yao & Chen, 2002) |
| LTriDP | Local Tri Directional Pattern(Verma & Raman, 2016) |
| Nanni et al. | Local binary patterns variants as texture descriptors (Nanni et al., 2010) |
| LBP | Local Binary Pattern(T Ojala et al., 2002) |
| LDGP | Local Directional Gradient Pattern(Chakraborty, Singh, & Chakraborty, 2017) |
| LNDP | Local Neighborhood Difference Pattern(Verma & Raman, 2017) |
| LEPSEG | Local Edge Pattern for Segmentation(Yao & Chen, 2002) |

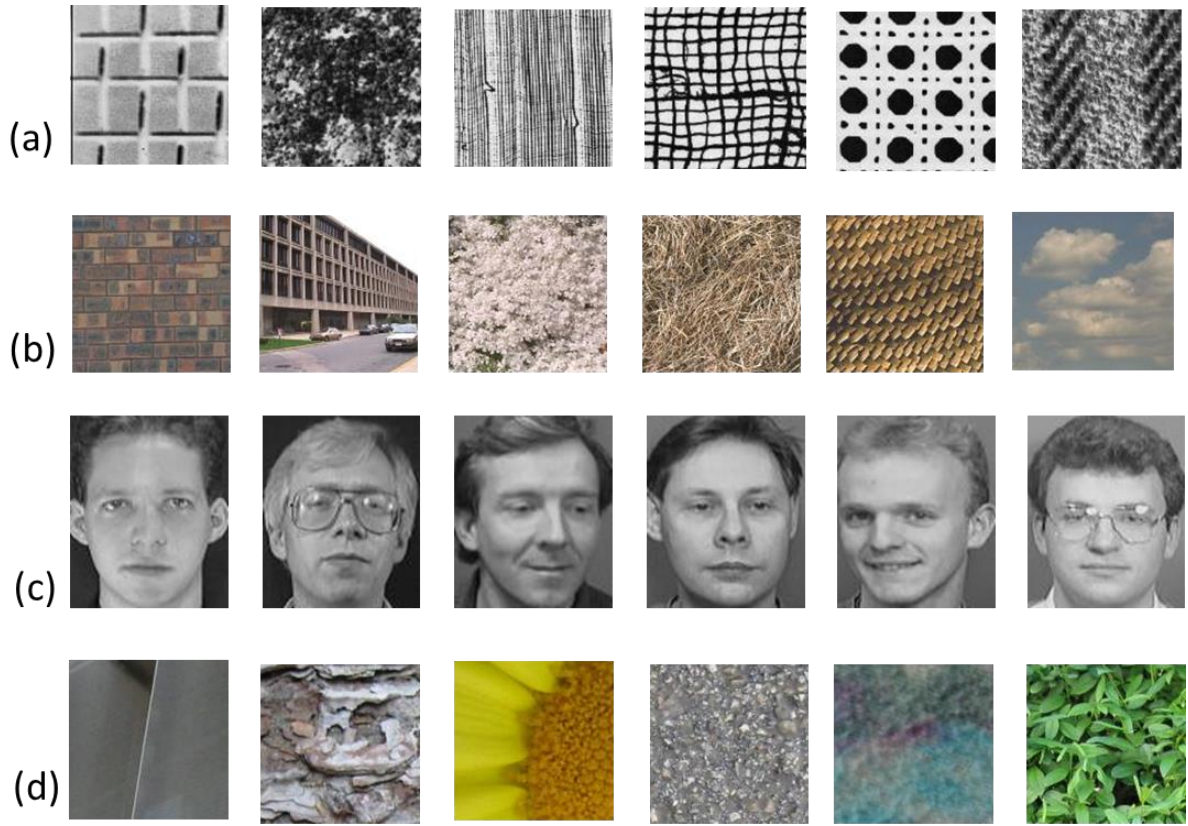

Fig. 6. Sample images from different datasets.

## 4.1 Dataset-1

The Brodatz texture dataset[1] contains 112 texture images each of size 640×640. The images have been divided into 25 sub images each of size 128×128. Thus there are 112 categories and each category contains 25 images. The results in terms of $P_{total}$ and $R_{total}$ obtained using our proposed method on this dataset have been presented with the help of graphical plots shown in Fig. 7(a) and Fig. 7(b). In our experiment, we have initially retrieved 25 images and then increased the number of retrieved images in steps of 5 images. We have retrieved a maximum of 70 images for this dataset. The performance of the proposed method is better than those obtained with state of the art methods such as CSLBP, LEPINV, LEPSEG, LBP, Nanni et al, LDGP, LNDP and LTriDP by 25.61%, 23.77%, 15.59%, 9.09%, 12.29%, 14.94%, 3.92 %  and 3.04% on Average Retrieval Rate as shown in Table 3.

---

[1] http://multibandtexture.recherche.usherbrooke.ca/original_brodatz.html (accessed on 17/08/2017)

Fig. 6(a) shows the images from the dataset. In Fig. 8, the first image of each row represents the query image and the remaining images show the retrieved images for each query image. In Fig. 9 we have shown the comparison between the performance of sign and magnitude pattern of our proposed method by calculating the precision and recall values considering each individually. As is evident from the graph the sign pattern performs better than the magnitude pattern.

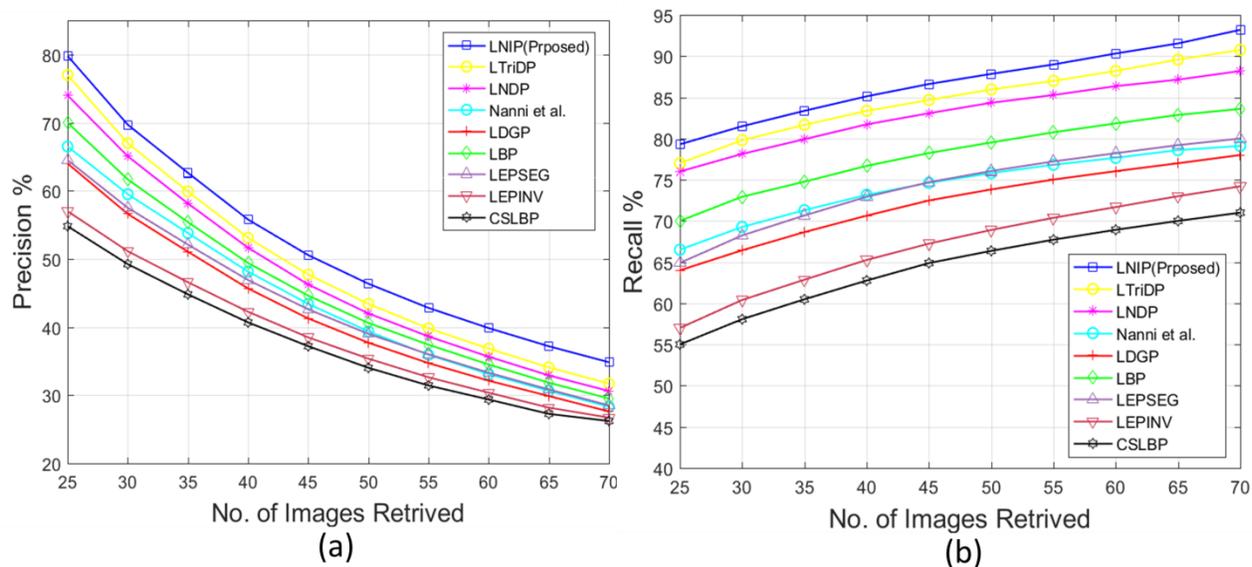

**Fig. 7. Precision and recall with number of images retrieved for Brodatz database.**

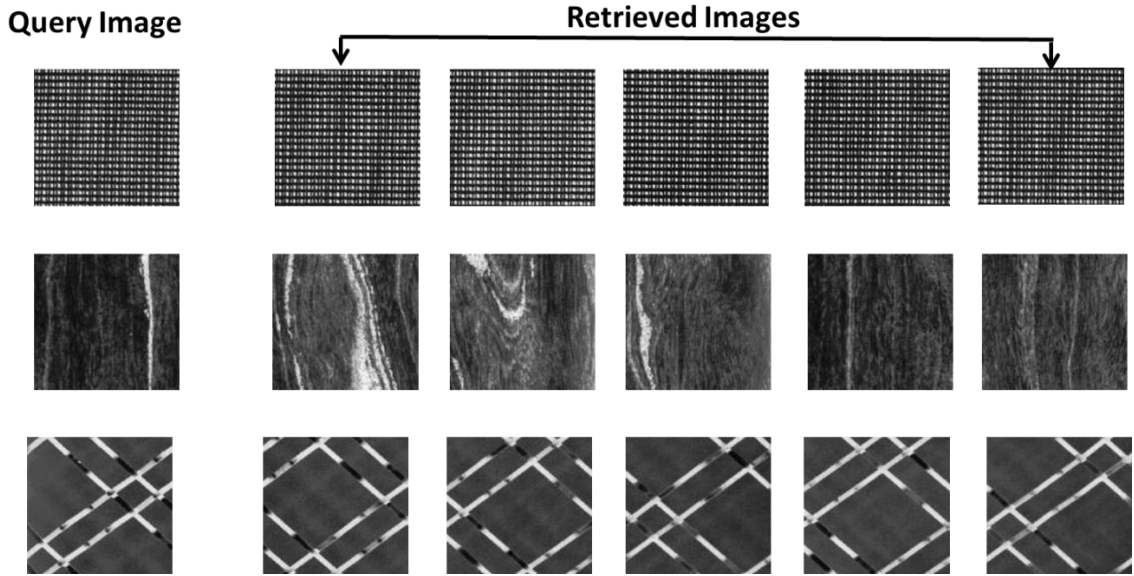

**Fig. 8. Query image and retrieved images from Brodatz database.**

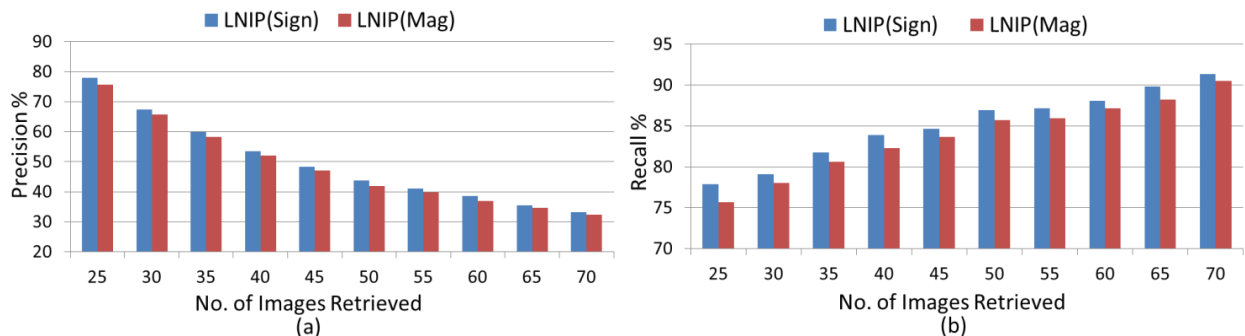

**Fig. 9. Precision and recall of proposed methods for Brodatz database.**

## 4.2 Dataset-2

For this experiment, the second database that we have used is MIT-Vistex[2] database. It contains 40 texture images each of size 512×512. For retrieval purpose, each image has been subdivided into images of size 128× 128. Thus, the dataset contains 40 different types of images with 16 images of each type. Similar to dataset 1, we have initially retrieved 16 images and then increased the number of retrieved images by 16. The maximum number of images retrieved is 96. We have calculated the Precision Rate and Recall Rate for all images in the database and compared it with the same set of methods compared in Fig. 7. Fig. 10 shows a graph in support of our observations. Fig. 6(b) shows some sample images from our dataset while Fig. 11 shows some query images and their corresponding retrieved images. The sign pattern of LNIP performs better than the magnitude pattern of LNIP as shown in Fig.12. The proposed method shows an improvement over CSLBP by 17.20%, LEPINV by 19.66%, LEPSEG by 11.67%, LBP by 7.54%, Nanni et al by 8.11 % LDGP by 9.90%, LNDP by 4.39% and LTriDP by 3.31 in terms of Average Retrieval Rate(ARR).

---
[2] MIT Vision and Modeling Group, Cambridge, Vision texture, available online: http://vismod.media.mit.edu/pub/.

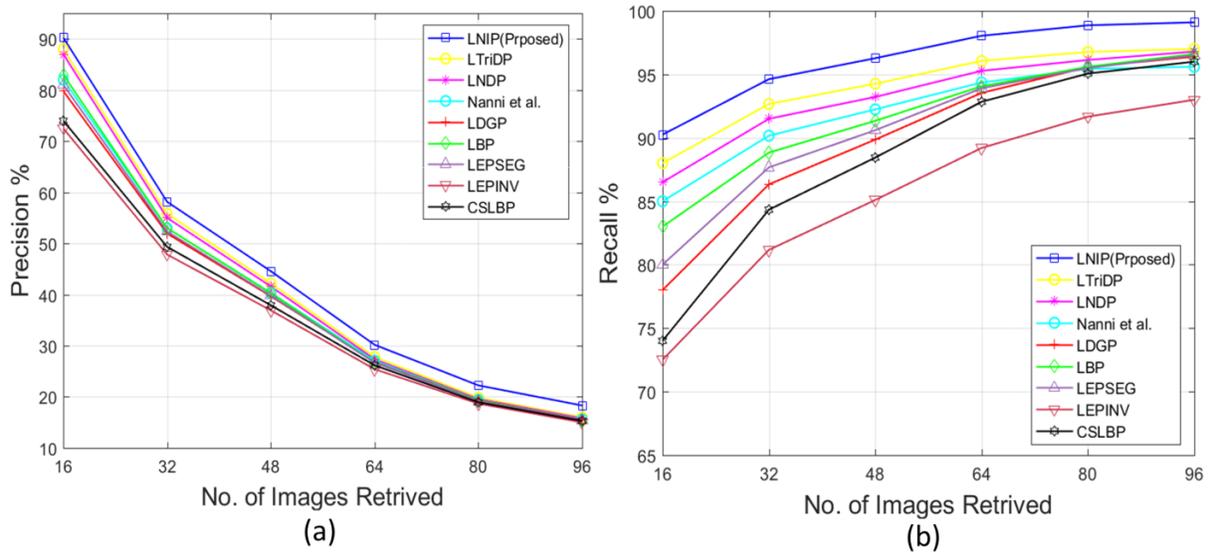

**Fig. 10. Precision and recall with varying number of images retrieved for database 2.**

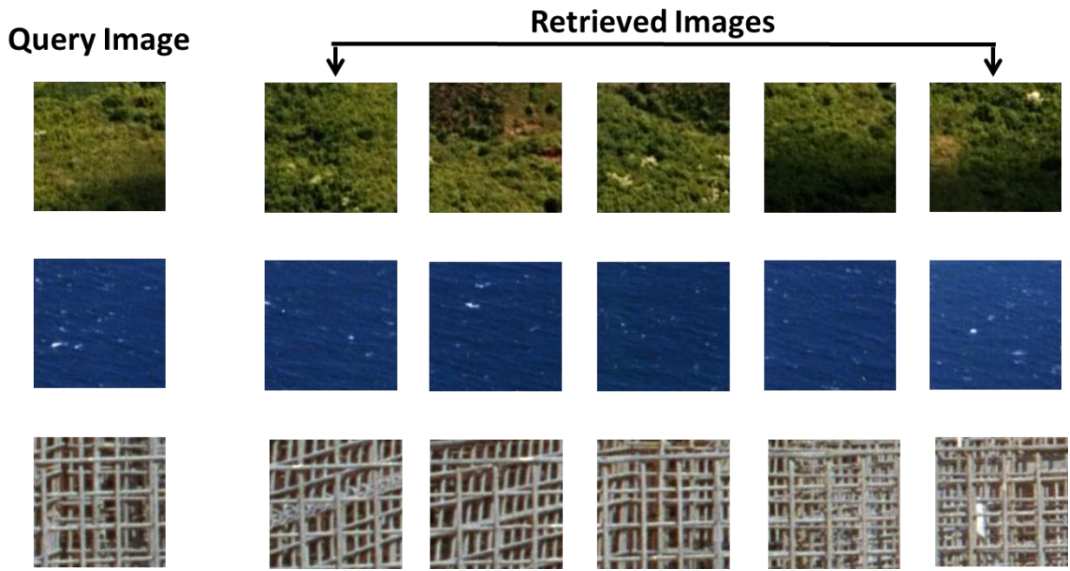

**Fig. 11. Query image and retrieved images from MIT-Vistex database.**

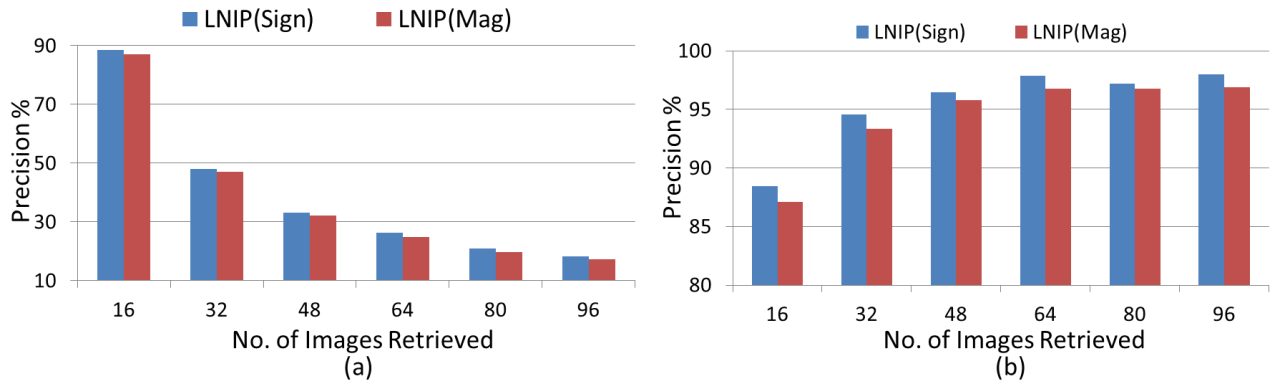

**Fig. 12. Precision and recall of proposed methods for MIT-Vistex database.**

### 4.3 Dataset-3

The third set of images considered for the comparative study of our method with several recently developed methods for image retrieval is the AT&T face dataset[3]. It contains images of 40 different types with 10 images of each type. Thus, there are a total of 400 images. The size of each image in the database is 92×112. Some sample images from this dataset are shown in Fig. 6(c). For this dataset, we have initially retrieved 1 image and then increased the number of retrieved images by 1. In this manner, we have retrieved a maximum of 10 images for this dataset. The query images and their corresponding retrieved images are shown in Fig. 14. The images were captured at different times and under varying illumination. The proposed method outperforms the existing methods like CSLBP, LEPINV, LEPSEG, LBP, Nanni et al., LDGP, LNDP and LTriDP by 13.73 %, 31.45%, 20.78%, 14.89%, 4.39%, 11.47%, 3.47%, 2.73% when performance is compared using Average Retrieval Rate as a metric. The precision and recall curves have been shown with the help of graphs in Fig. 13(b) and 13(a). The comparative performance of sign and magnitude patterns of LNIP for this database is shown in Fig.15 and the observations are similar to the one in Fig. 12.

---

[3]The AT&T database of faces: http://www.uk.research.att.com/facedatabase.html, 2002.

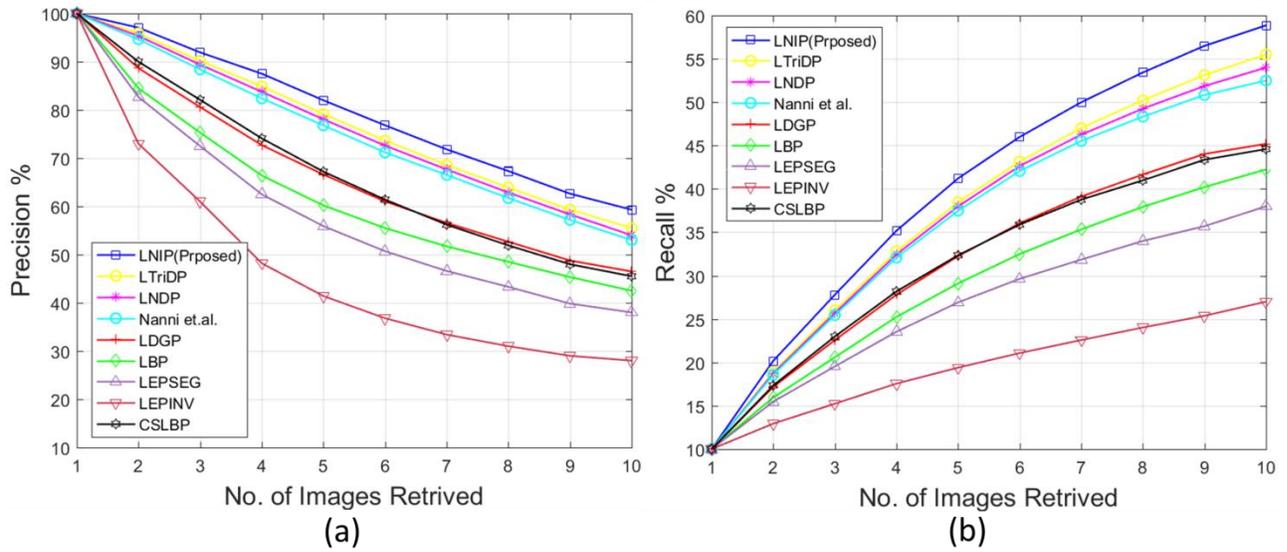

**Fig. 13. Precision and recall with varying number of images retrieved for AT&T face dataset.**

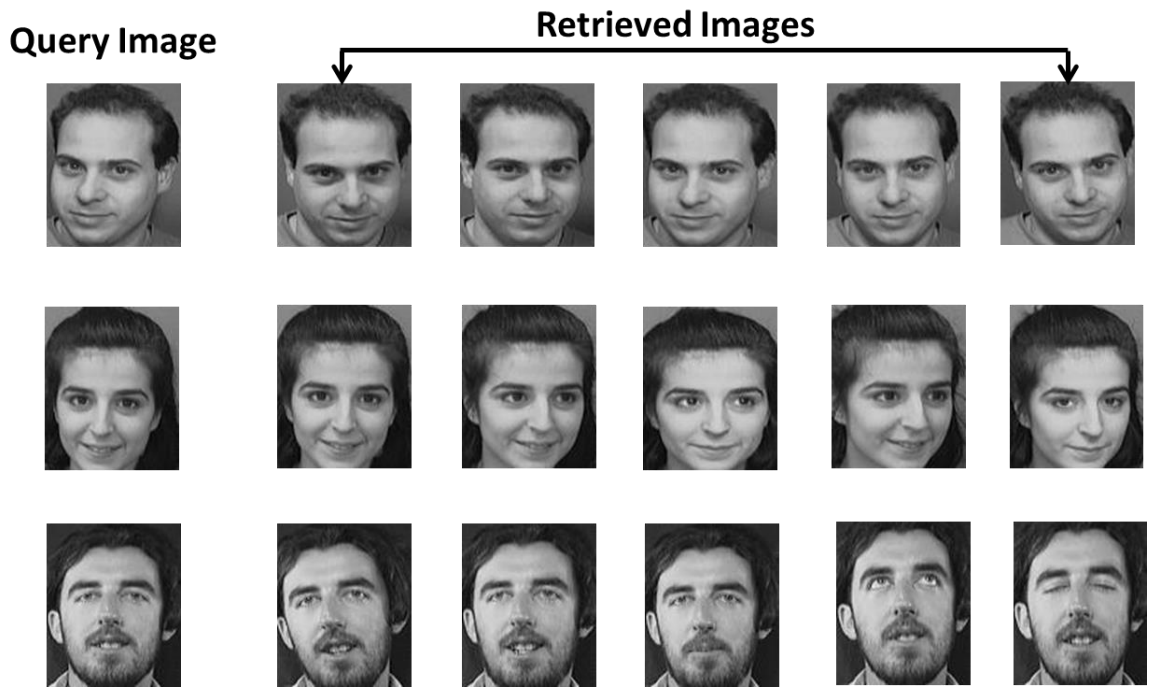

**Fig. 14. Query image and retrieved images from AT&T database.**

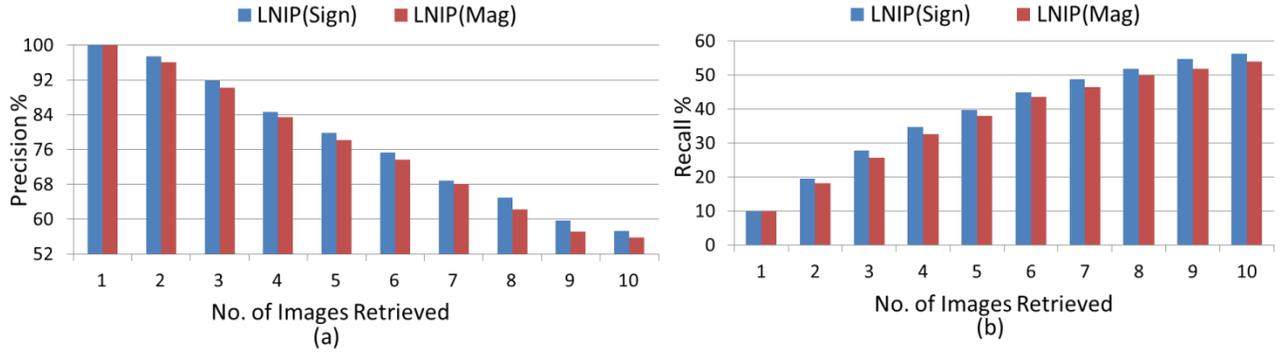

**Fig. 15. Precision and recall of proposed methods for AT&T database.**

### 4.4 Dataset-4

The Salzburg[4] texture Database contains images of size 128×128. There are a total of 7616 images. There are a total of 476 categories and each category contains 16 images. Different types of textures like wood, rubber, etc. are presented in the database. Sample images from the database are presented in Fig. 6(d). Each image of the database is treated as a query image. Fig. 17 shows some query images and their corresponding retrieved image. The number of images retrieved for each category for this experiment is initially considered as 16. This is increased in small steps of 16 images. The maximum number of images for this dataset retrieved in our experiment is 112. The proposed method clearly shows an improvement on Average retrieval rate (ARR) over the recently developed methods for content based image retrieval such as CSLBP by 22.99%, LEPINV by 33.41%, LEPSEG by 24.97%, LBP by 6.29%, Nanni et al. by 8.67%, LDGP by 9.86%, LNDP by 4.13%, LTriDP by 2.96 %. The precision and recall curves are shown in Fig. 16(a) and Fig. 16(b). The comparative performance of sign and magnitude patterns of LANIP for this dataset is shown in Fig. 18(a) and Fig. 18(b).

---

[4] http://www.wavelab.at/sources/STex/

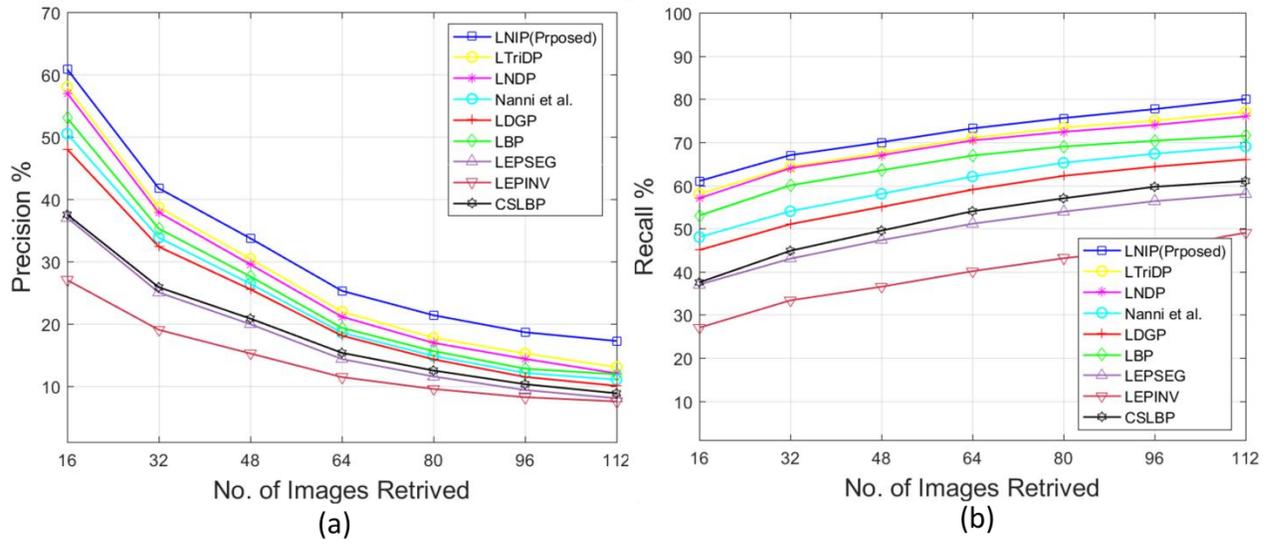

**Fig. 16. Precision and recall with varying number of images retrieved for Stex database.**

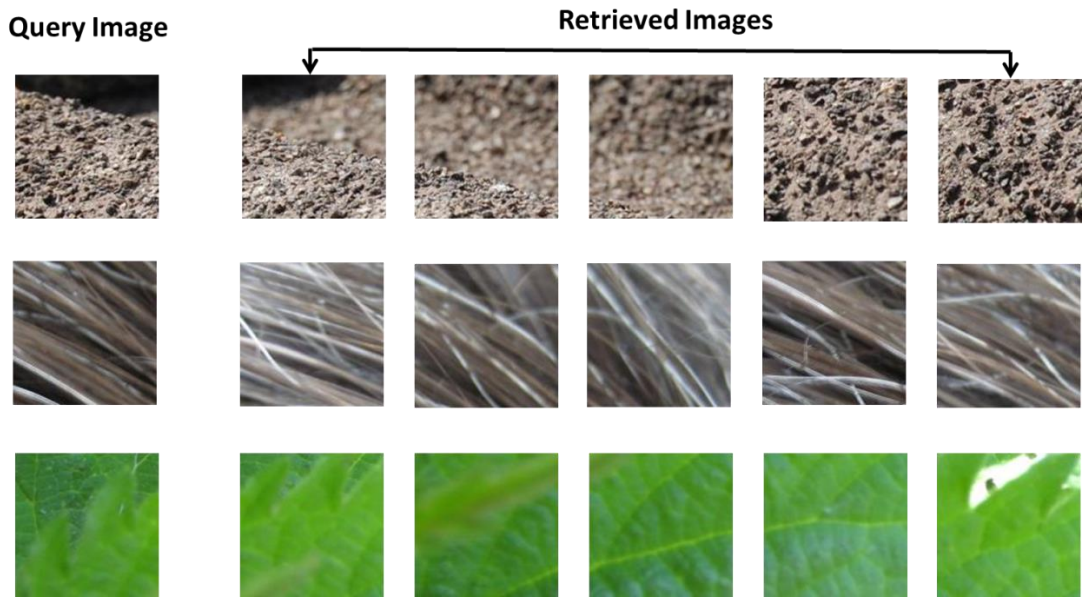

**Fig. 17. Query image and retrieved images from Stex database.**

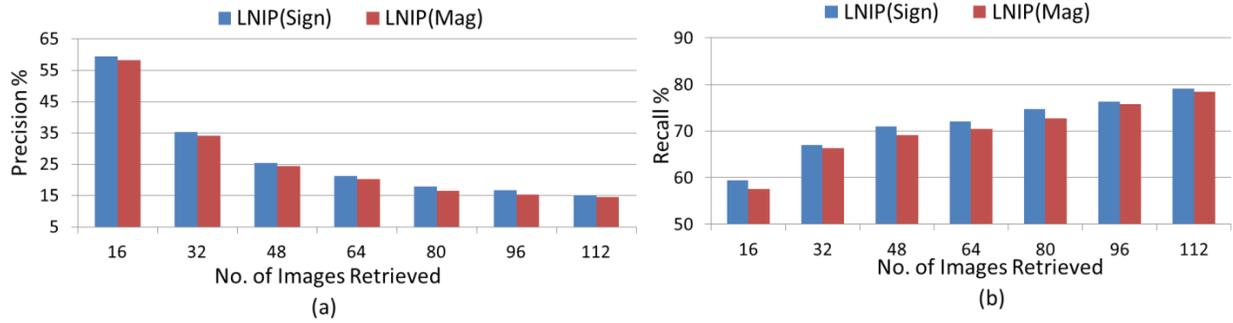
**Fig. 18. Precision and recall of proposed methods for Stex database.**

A comparative study of feature vector length, feature extraction time and image retrieval time has been presented in Table 4. The feature vector length of the proposed method is 512 which is higher than that of LBP but lower than most of the recent patterns like LTriDP, and LTrP. The feature extraction and image retrieval have been presented in table. As is evident from the table, the feature extraction and image retrieval time is comparable with recent patterns developed for image retrieval. As mentioned earlier, we have evaluated the performance of the proposed method on different datasets by considering various similarity metrics. The results are shown in Table 5.

**Table 3: Average Retrieval Rate for different datasets**

| Different Feature | Brodatz | MIT-Vistex | AT&T | STex |
|---|---|---|---|---|
| CSLBP | 53.54 | 72.61 | 43.29 | 36.22 |
| LEPINV | 55.38 | 70.15 | 25.57 | 25.80 |
| LEPSEG | 63.56 | 78.14 | 36.24 | 34.24 |
| LBP | 70.06 | 82.27 | 42.13 | 52.92 |
| Nanni et al. | 66.86 | 81.70 | 52.63 | 50.54 |
| LDGP | 64.21 | 79.91 | 45.55 | 49.35 |
| LNDP | 75.23 | 85.42 | 53.55 | 55.08 |
| LTriDP | 76.11 | 86.50 | 54.29 | 56.25 |
| LNIP(Proposed) | 79.15 | 89.81 | 57.02 | 59.21 |

**Table 4: Runtime performance and feature length using different methods**

| | Feature Length | Feature Extraction Time | Retrieval Time |
|---|---|---|---|
| CSLBP | 16 | 0.0196 | 0.0322 |
| LEPINV | 72 | 0.0720 | 0.0323 |
| LEPSEG | 512 | 0.0371 | 0.0330 |
| LBP | 256 | 0.0192 | 0.0325 |
| Nanni et al. | 93 | 0.0380 | 0.0340 |

| LDGP | 64 | 0.0370 | 0.0329 |
| LTriDP | 768 | 0.0305 | 0.0332 |
| Proposed | 512 | 0.0348 | 0.0345 |

**Table 5: Comparative study with different distance matrices**

|  | Brodatz | MIT-Vistex | AT&T | STex |
|---|---|---|---|---|
| d1 | 79.15 | 89.81 | 57.02 | 59.21 |
| Euclidean | 67.47 | 79.50 | 45.30 | 47.69 |
| Manhattan | 75.29 | 85.09 | 52.11 | 54.58 |
| Canberra | 63.10 | 85.62 | 52.36 | 43.12 |
| Chi Square | 76.10 | 86.21 | 54.83 | 56.61 |

From the experiments, it is very much evident that our proposed method outperform most of the state-of-the-arts texture descriptor in a clear margin. The major reason of this improved performance is that we have considered the mutual relationship among adjacent neighbors, while most of the existing methods do not take that into account. Also, one more important thing to notice that concatenation of sign pattern ($LNIP_S$) and magnitude pattern ($LNIP_M$) improve the performance significantly because of their complementary information.

## 5. Conclusion

In this paper, a new feature named Local Neighborhood Intensity Pattern has been developed for image retrieval and abbreviated as LNIP. It has been separated into sign and magnitude components $LNIP_S$ and $LNIP_M$ respectively. Unlike previously developed approaches for image retrieval which are based on comparing the intensity of a pixel with the center pixel in a 3×3 neighborhood, the proposed method takes into account the relative intensity difference of a particular pixel with its adjacent neighbors for pattern calculation and provides robustness against illumination and gray level changes. The method has been tested on standard databases and compared with existing techniques by calculating the precision and recall values for them. The proposed method justifies its superiority over existing approaches in terms of both precision and recall.

The proposed expert image retrieval system is able to produce a desired accuracy level but in order to achieve a high precision and recall value we had to concatenate two patterns namely sign (LNIP$_S$) and magnitude (LNIP$_M$) into a single feature descriptor as most of the previous methods did not consider both of them simultaneously, but as a result the feature vector length of our method has been 512. We are looking forward to devise feature descriptors merging these two patterns into one single descriptor that will serve both the purposes of sign and magnitude together and thus the feature vector length can be minimized considerably. Again, due to two patterns the feature extraction and image retrieval time is somewhat high but the single descriptor serving both sign and magnitude information simultaneously can reduce it considerably. This could be our future work of concern in modifying our paper. Further, this work can be extended to multi-scale feature extraction for image retrieval by integrating with Gabor transform or Gaussian filters. Overall, as our work achieved good accuracy level so the technique can be adapted for efficient image retrieval in real time systems.